%% file: nips_2016.tex
\documentclass{article}

\usepackage[final, nonatbib]{nips_2016}
\input{defns2}

\usepackage[utf8]{inputenc} 
\usepackage[T1]{fontenc}    
\usepackage{hyperref}       
\usepackage{url}            
\usepackage{booktabs}       
\usepackage{amsfonts}       
\usepackage{nicefrac}       
\usepackage{microtype}      
\usepackage{graphicx}
\usepackage{float}
\usepackage{amsmath}

\title{Normalizing Flows on Riemannian Manifolds}

\author{
   Mevlana C. Gemici\\
   Google DeepMind \\
   \texttt{mevlana@google.com} \\
   \And
   Danilo J. Rezende\\
   Google DeepMind \\
   \texttt{danilor@google.com} \\
   \And
   Shakir Mohamed \\
   Google DeepMind \\
   \texttt{shakir@google.com} \\
}

\begin{document}
\maketitle
\vspace{-0.1in}
\begin{abstract}
We consider the problem of density estimation on Riemannian manifolds.
Density estimation on manifolds has many applications in fluid-mechanics, optics and plasma physics and it appears often when dealing with angular variables (such as used in protein folding, robot limbs, gene-expression) and in general directional statistics. In spite of the multitude of algorithms available for density estimation in the Euclidean spaces $\vR^n$ that scale to large n (e.g. normalizing flows, kernel methods and variational approximations), most of these methods are not immediately suitable for density estimation in more general Riemannian manifolds.
We revisit techniques related to homeomorphisms from differential geometry for projecting densities to sub-manifolds and use it to generalize the idea of normalizing flows to more general Riemannian manifolds. The resulting algorithm is scalable, simple to implement and suitable for use with automatic differentiation. We demonstrate concrete examples of this method on the n-sphere $\vS^n$.
\end{abstract}

\vspace{-0.1in}
In recent years, there has been much interest in applying variational inference techniques to learning large scale probabilistic models in various domains, such as images and text \cite{rezende2014stochastic, kingma2014stochastic, gregor2015draw, eslami2016attend, rezende2016one,JMLR:v14:hoffman13a}. One of the main issues in variational inference is finding the best approximation to an intractable posterior distribution of interest by searching through a class of known probability distributions. 
The class of approximations used is often limited, e.g., mean-field approximations, implying that no solution is ever able to resemble the true posterior distribution. This is a widely raised objection to variational methods, in that unlike MCMC, the true posterior distribution may not be recovered even in the asymptotic regime.
To address this problem, recent work on Normalizing Flows \cite{rezende2015variational}, Inverse Autoregressive Flows \cite{inverseAutoregressive}, and others \cite{realnvp, salimansHamiltonian} (referred collectively as normalizing flows), focused on developing scalable methods of constructing arbitrarily complex and flexible approximate posteriors from simple distributions using transformations parameterized by neural networks, which gives these models universal approximation capability in the asymptotic regime. In all of these works, the distributions of interest are restricted to be defined over high dimensional Euclidean spaces. 

There are many other distributions defined over special homeomorphisms of Euclidean spaces that are of interest in statistics, such as Beta and Dirichlet (n-Simplex); Norm-Truncated Gaussian (n-Ball); Wrapped Cauchy and Von-Misses Fisher (n-Sphere), which find little applicability in variational inference with large scale probabilistic models due to the limitations related to density complexity and gradient computation \cite{ClusteringHypersphere,VMFClustering,UlrichGeodesics,GenerativeDirectional}. Many such distributions are unimodal and generating complicated distributions from them would require creating mixture densities or using auxiliary random variables. Mixture methods require further knowledge or tuning, e.g. number of mixture components necessary, and a heavy computational burden on the gradient computation in general, e.g. with quantile functions \cite{alexMixtureWork}. Further, mode complexity increases only linearly with mixtures as opposed to exponential increase with normalizing flows. Conditioning on auxiliary variables \cite{auxiliaryVar} on the other hand constrains the use of the created distribution, due to the need for integrating out the auxiliary factors in certain scenarios. In all of these methods, computation of low-variance gradients is difficult due to the fact that simulation of random variables cannot be in general reparameterized (e.g. rejection sampling \cite{rejectionSampledGradient}).
In this work, we present methods that generalizes previous work on improving variational inference in $\vR^n$ using normalizing flows to Riemannian manifolds of interest such as spheres $\vS^n$, tori $\vT^n$ and their product topologies with $\vR^n$, like infinite cylinders. 

\begin{figure}[H]
  \centering
  \includegraphics[height=0.25\linewidth]{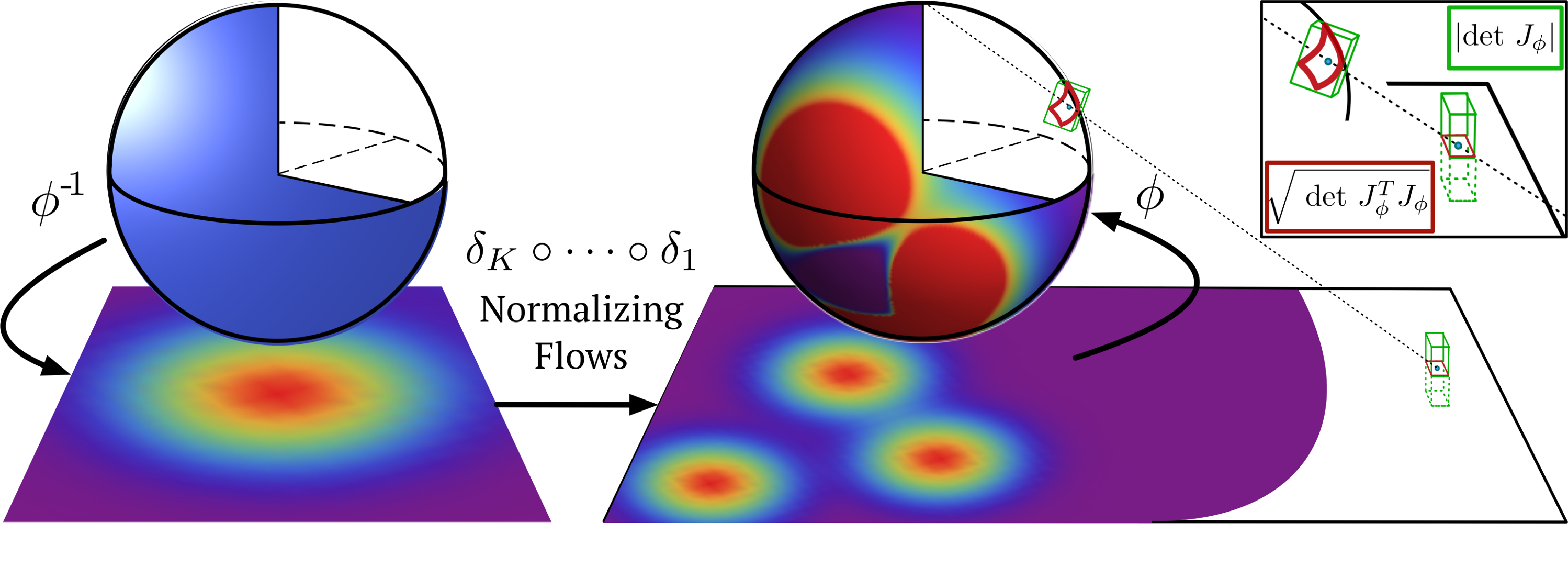}
  \includegraphics[height=0.25\linewidth]{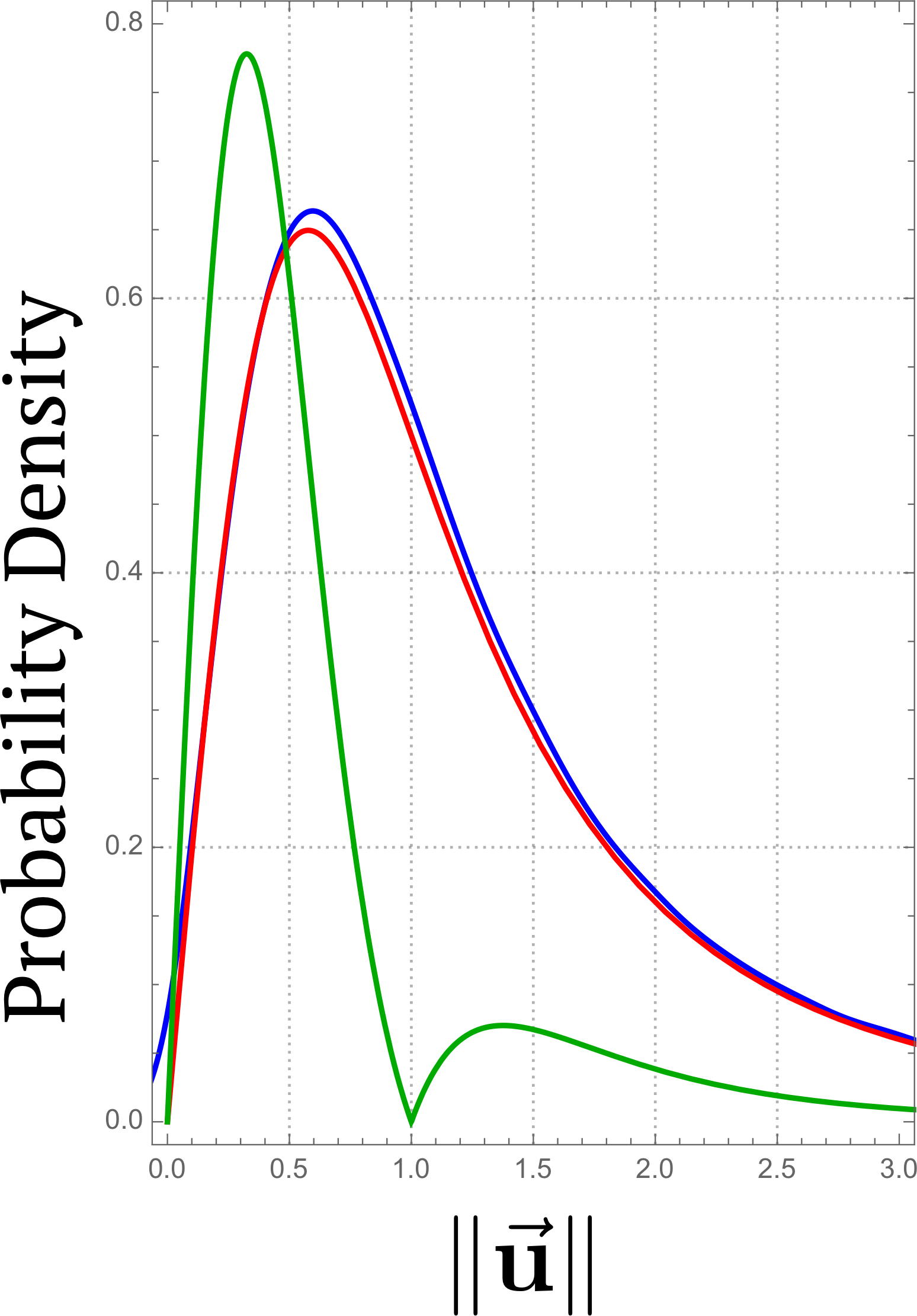}
  \caption{Left: Construction of a complex density on $\vS^n$ by first projecting the manifold to $\vR^{n}$, transforming the density and projecting it back to $\vS^n$. Right: Illustration of transformed ($\vS^2 \rightarrow \vR^2$) densities  corresponding to an uniform density on the sphere. Blue: empirical density (obtained by Monte Carlo); Red: Analytical density from equation \eqref{eq.sphere}; Green: Density computed ignoring the intrinsic dimensionality of $\vS^n$.}
  \vspace{-.2in}
  \label{fig1}
\end{figure}

These special manifolds $\vM \subset \vR^m$ are homeomorphic to the Euclidean space $\vR^n$ where $n$ corresponds to the dimensionality of the tangent space of $\vM$ at each point. A homeomorphism is a continuous function between topological spaces with a continuous inverse (bijective and bicontinuous). It maps point in one space to the other in a unique and continuous manner. An example manifold is the unit 2-sphere, the surface of a unit ball, which is embedded in $\vR^3$ and homeomorphic to $\vR^2$ (see Figure \ref{fig1}). 

In normalizing flows, the main result of differential geometry that is used for computing the density updates is given by, $\vd\vec{\vx} = |\text{det } J_{\vphi}|\text{ } \vd\vec{\vu} $ and represents the relationship between differentials (infinitesimal volumes) between two equidimensional Euclidean spaces using the Jacobian of the function $\vphi:\vR^n \rightarrow \vR^n$ that transforms one space to the other. This result only applies to transforms that preserve the dimensionality. 
However, transforms that map an embedded manifold to its intrinsic Euclidean space, do not preserve the dimensionality of the points and the result above become obsolete. Jacobian of such transforms $\vphi:\vR^n \rightarrow \vR^m$ with $m>n$ are rectangular and an infinitesimal cube on $\vR^n$ maps to an infinitesimal parallelepiped on the manifold. The relation between these volumes is given by $\vd\vec{\vx} = \sqrt{\text{ }\text{det } \vG } \text{ } \vd\vec{\vu}$, where $\vG =J_{\vphi}^TJ_{\vphi}$ is the metric induced by the embedding $\vphi$ on the tangent space $T_x\vM$, \cite{ben1999change, ben2000application, berger2012differential}. The correct formula for computing the density over $\vM$ now becomes
:

\begin{align}
\int_{\vM \subset \vR^m} \vf(\vec{\vx}) \vd\vec{\vx} = \int_{\vR^n} (\vf \circ \vphi)(\vec{\vu} )\text{ }\sqrt{\text{ }\text{det } \vG }\text{ } \vd\vec{\vu} = \int_{\vR^n} (\vf \circ \vphi)(\vec{\vu} )\left(\sqrt{\text{ }\text{det } J_{\vphi}^TJ_{\vphi}} \text{ }\right)\vd\vec{\vu} 
\end{align}

The density update going from the manifold to the Euclidian space, $\vec{\vx}\in\vS^n \rightarrow \vec{\vu}\in\vR^n$, is then given by:
\begin{align}
\vp(\vec{\vu}) = (\vf \circ \vphi)(\vec{\vu} ) \sqrt{\text{ }\text{det } J_{\vphi}^TJ_{\vphi}(\vec{\vu})} \text{ } = \vf(\vec{\vx}) \sqrt{\text{ }\text{det } J_{\vphi}^TJ_{\vphi}(\vphi^{-1}(\vec{\vx}))} \text{ }
\end{align}
As an application of this method on the $n$-sphere $\vS^n$, we introduce \emph{Inverse Stereographic Transform} and define it as: 
$\vphi(\vu) : \vR^n\rightarrow \vS^n\subset\vR^{n+1}$,
\begin{align}
\vec{\vx} = \vphi(\vec{\vu}) =
\left[
\begin{array}{c}
 2\vu/(\vu^T\vu+1) \\
 1-2/(\vu^T\vu+1) \\
\end{array} 
\right]
\end{align}
which maps $\vR^n$ to $\vS^n$ in a bijective and bicontinuous manner. The determinant of the metric $\vG(\vx)$ associated with this transformation is given by: 
\begin{align}
\text{det } \vG = \text{det }{{J_{\vphi}}(\vx)}^T{J_{\vphi}}(\vx)=
\text{}
\left(\frac{2}{\vx^T\vx+1}\right)^{2n}\label{eq.sphere}
\end{align}
Using these formulae, on the left side of Figure 1, we map a uniform density on $\vS^2$ to $\vR^2$, enrich this density, using e.g. normalizing flows, and then map it back onto $\vS^2$ to obtain a multi-modal (or arbitrarily complex) density on the original sphere. On the right side of Figure 1, we show that the density update based on the Riemannian metric, i.e. $\sqrt{\text{ }\text{det } J_{\vphi}^TJ_{\vphi}} \text{ }$ (red), is correct and closely follows the kernel density estimate based on 500k samples (blue). We also show that using the generic volume transformation formulation for dimensionality preserving transforms, i.e. $|\text{det } J_{\vphi}|\text{ }$ (green), leads to an erroneous density and do not resemble the empirical distributions of samples after the transformation. 
\vfill


\input{nips_2016.bbl}
\end{document}

%% file: defns2.tex
\usepackage{amsmath,amsthm,amssymb}

\newcommand\cut[1]{}

\newcommand{\squishlist}{
   \begin{list}{$\bullet$}
    { \setlength{\itemsep}{0pt}      \setlength{\parsep}{3pt}
      \setlength{\topsep}{3pt}       \setlength{\partopsep}{0pt}
      \setlength{\leftmargin}{1.5em} \setlength{\labelwidth}{1em}
      \setlength{\labelsep}{0.5em} } }

\newcommand{\squishlisttwo}{
   \begin{list}{$\bullet$}
    { \setlength{\itemsep}{0pt}    \setlength{\parsep}{0pt}
      \setlength{\topsep}{0pt}     \setlength{\partopsep}{0pt}
      \setlength{\leftmargin}{2em} \setlength{\labelwidth}{1.5em}
      \setlength{\labelsep}{0.5em} } }

\newcommand{\squishend}{
    \end{list}  }

\newcommand{\myvec}[1]{\mathbf{#1}}
\newcommand{\myvecsym}[1]{\boldsymbol{#1}}

\newcommand{\vphi}{\myvecsym{\phi}}

\newcommand{\vd}{\myvec{d}}

\newcommand{\vf}{\myvec{f}}

\newcommand{\vp}{\myvec{p}}

\newcommand{\vu}{\myvec{u}}

\newcommand{\vx}{\myvec{x}}

\newcommand{\vG}{\myvec{G}}

\newcommand{\vM}{\myvec{M}}

\newcommand{\vR}{\myvec{R}}
\newcommand{\vS}{\myvec{S}}
\newcommand{\vT}{\myvec{T}}

\newcommand{\be}{\begin{equation}}
\newcommand{\ee}{\end{equation}}
\newcommand{\bea}{\begin{eqnarray}}
\newcommand{\eea}{\end{eqnarray}}
\newcommand{\beaa}{\begin{eqnarray*}}
\newcommand{\eeaa}{\end{eqnarray*}}

\DeclareMathAlphabet{\mathpzc}{OT1}{pzc}{m}{n}